\documentclass[10pt,twocolumn,letterpaper,dvipdfmx]{article}

\usepackage{cvpr}              

\usepackage{graphicx}
\usepackage{amsmath}
\usepackage{amssymb}
\usepackage{booktabs}
\usepackage{comment}

\usepackage[pagebackref,breaklinks,colorlinks]{hyperref}

\usepackage[capitalize]{cleveref}
\crefname{section}{Sec.}{Secs.}
\Crefname{section}{Section}{Sections}
\Crefname{table}{Table}{Tables}
\crefname{table}{Tab.}{Tabs.}

\def\cvprPaperID{*****} 
\def\confName{CVPR}
\def\confYear{2022}

\begin{document}

\title{5th Place Solution to Kaggle Google Universal Image Embedding Competition}

\author{Shingo Yokoi, \quad Noriaki Ota, \quad Shinsuke Yamaoka\\
NS Solutions Corp.\\
\{yokoi.shingo.a6q, ota.noriaki.4qp, yamaoka.shinsuke.5ke\}@jp.nssol.nipponsteel.com\\
}
\maketitle

\begin{abstract}
In this paper, we present our solution, which placed 5th in the kaggle Google Universal Image Embedding Competition in 2022.
We use the ViT-H visual encoder of CLIP from the openclip repository as a backbone and train a head model composed of BatchNormalization and Linear layers using ArcFace. The dataset used was a subset of products10K, GLDv2, GPR1200, and Food101. And applying TTA for part of images also improves the score.
With this method, we achieve a score of 0.684 on the public and 0.688 on the private leaderboard. Our code is available. \url{https://github.com/riron1206/kaggle-Google-Universal-Image-Embedding-Competition-5th-Place-Solution}

\end{abstract}

\section{Introduction}
\label{sec:intro}

Google Universal Image Embedding Competition~\cite{complink} is a multi-domain image retrieval task that identifies images that have the same instance as the query image from a set of images from multiple domains (e.g., clothing, packaged goods, landmarks).
Previously, the Google Landmark Challenges~\cite{landmarkreclink}~\cite{landmarkretlink} focused on instance-level recognition (ILR) of a single domain, and this competition has been extended to multiple domains.
Our approach used the ViT-H~\cite{dosovitskiy2021an} visual encoder of the CLIP~\cite{DBLP:journals/corr/abs-2103-00020} model trained on the LAION-2B~\cite{schuhmann2022laionb} dataset as the backbone. CLIP is well suited for multidomain tasks such as this one because it has such a high general recognition capability that it can recognize many classes with high accuracy even in a zero-shot task.
In this competition, the size of the output embedding vector is limited to 64 dimensions, so compression of the dimension is necessary. Therefore, we used ArcFace~\cite{DBLP:journals/corr/abs-1801-07698} to train a projection head composed of two layers, BatchNormalization and Linear. this architecture is based on MAE~\cite{DBLP:journals/corr/abs-2111-06377}. The dataset used is the subset of products10K~\cite{DBLP:journals/corr/abs-2008-10545}, GLDv2~\cite{DBLP:journals/corr/abs-2004-01804}, GPR1200~\cite{DBLP:journals/corr/abs-2111-13122}, and Food101~\cite{bossard14}. And applying TTA also improves the score.
As a result, we achieved 0.688 mAP@5 on the private leaderboard, achieving 5th place.
\section{Method}
\subsection{Architecture}
\begin{figure*}[t]
\begin{center}
   \includegraphics[width=1.0\linewidth]{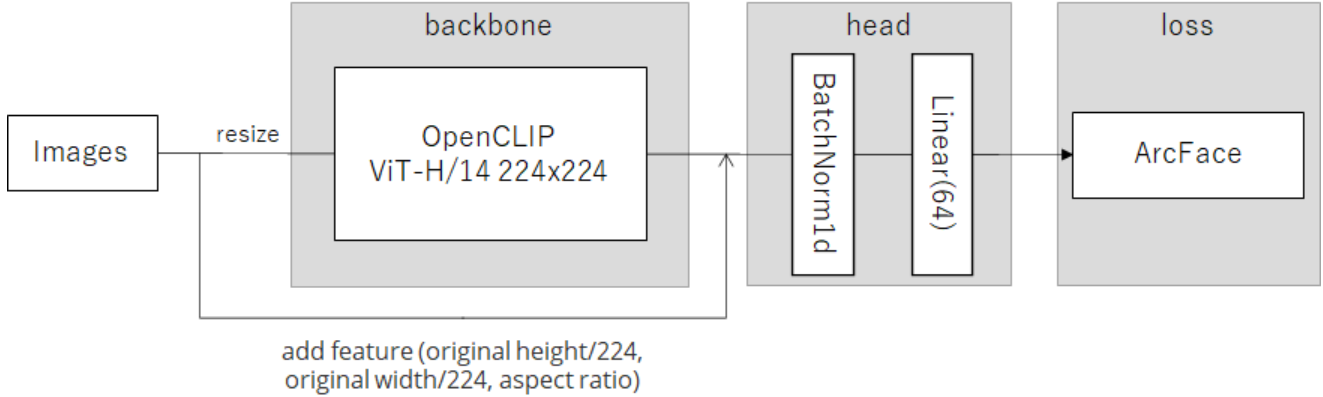}
\end{center}
   \caption{Model architecture.}
   \label{fig:modelarc}
\end{figure*}

The architecture we used is shown in \cref{fig:modelarc}.
We used  the ViT-H visual encoder of CLIP model trained on the LAION-2B dataset as a backbone. This model was released during the competition and is used by most of the top teams.
We tried many other models, but no one had a better performance than this model. We think the key point is that it is a large model trained on a larger dataset.
The input size of the image for this model is 224x224 (RGB) and the output is a 1024-dimensional vector.
The head architecture is based on MAE and is composed of two layers: BatchNormalization(affine=False) and Linear.
And head receives a 1027-dimensional vector that is a concatenation of the 1024-dimensional vector that is the output of ViT-H and the original height \// 224, original width \// 224, and aspect ratio of the input image.

\subsection{Training \& Validation}
We used ArcFace (scale=30.0, margin=0.5) for training of head. We tried various other similar methods (MSLoss~\cite{Wang2019MultiSimilarityLW}, ContrastiveLoss~\cite{1640964}, ProxyAnchorLoss~\cite{DBLP:journals/corr/abs-2003-13911}, AdaFace~\cite{kim2022adaface}, CircleLoss~\cite{9156774}, NTXentLoss~\cite{DBLP:journals/corr/abs-1807-03748}, SubCenterArcFaceLoss~\cite{conf/eccv/DengGLGZ20}), but ArcFace performed the best.
We used SAM~\cite{DBLP:journals/corr/abs-2010-01412}+AdamW (weight\_decay=0.1) for Optimizer. By using a very large weight decay, we can reduce the overfit to the train data and improve the score.
For the Learning Rate scheduler, CosineAnnealing (1e-2 to 1e-4, with 1e-4 to 1e-2 3epochs warmup) is used, the batch size is 256, and the epoch is 1000.
Also, when training the head, pre-computing and storing the ViT-H outputs made it possible to train at a very fast speed. This can be trained in about 10 hours using the NVIDIA RTX3090.
To check the performance of zero-shot, we prepared a test set with only data from classes not existing in train and validation. For submission, we used weights with high MAP@5 scores in the test set. However, this score has a low correlation with the LB score, and we finally considered a high LB score to be the most important.

\subsection{Dataset}
We used subsets of products10K, GLDv2, GPR1200, and Food101 for training. The number of classes and samples from each dataset is shown in  \cref{tab:dataset}.
\begin{table}[b]
  \centering
  \begin{tabular}{crr}
    \toprule
    Dataset & Classes & Samples \\
    \midrule
    GLDv2 & 5000 & 249350 \\
    Products10k & 9691 & 141931 \\
    GPR1200\_landmark & 200 & 2000 \\
    GPR1200\_face & 200 & 2000 \\
    GPR1200\_instre & 200 & 2000 \\
    GPR1200\_sketch & 200 & 2000 \\
    GPR1200\_sop & 200 & 2000 \\
    Food101 & 101 & 3030 \\
    \bottomrule
  \end{tabular}
  \caption{Number of classes and samples from each datasets.}
  \label{tab:dataset}
\end{table}
We used the clean version of the GLDv2 dataset. This dataset has a very large number of data, so we randomly selected 5000 classes from the classes that have more than 49 samples. All the data from the Products10K dataset were used. The GPR1200 dataset contains 200 classes of data each sampled from six different datasets. We used them all except that from the iNaturalist. The Food101 dataset was randomly removed and the maximum number of samples for each class was 30. We tried many other datasets but did not use them due to lower scores on the public leaderboard. We also tried adjusting the number of data according to the distribution of the types of objects in the test set revealed by the host, but this did not work well.

\subsection{Inference tricks}
We improved scores by preprocessing the input images and by using TTA (Test Time Augmentation).
There are two key points to input image preprocessing: first, it is important to use the same resizing method used when training the ViT-H encoder of CLIP. This means using antialias and BICUBIC for interpolation. The second is to pad the image to a square and then resize it to avoid changing the aspect ratio.
Next, when applying TTA, transforms such as flip were not effective, but such as crop or aspect ratio changes were effective. However, because we used a large encoder, we did not have enough time to apply TTA to all images. Therefore, we decided to apply TTA only to the no-square images. And we also applied TTA three times to images predicted to be apparel, packaged, or toy using the classification head trained using the 130K images dataset~\cite{130kimages}.
Finally, applying a slight center crop improved the scores. This is because important objects are often located in the center of the image, and center cropping has the same effect as increasing the resolution of the images. However, for example, in landmark images, information around objects is often important as well, so excessive center cropping can worsen the score.

\section{Findings}
We used the ViT-H encoder and were able to make a competitive model by training only a head. However, we found that the encoder also needs to be finetuned to make a better model. Team cuilab.ai has shown us a great solution to this~\cite{cuisol}. We also tried the approach of training with a smaller LR to avoid over-updating the encoder parameters, but it did not improve the score. The next time a similar competition is held, it will be a highly competitive battle.

\section{Conclusion}
In this paper, we presented our solution to the Google Universal Image Embedding competition. We use the ViT-H visual encoder of CLIP from the openclip repository as a backbone and train a head model composed of BatchNormalization and Linear layers using ArcFace. Tuning the dataset and using some tricks at inference, we reached a final score of 0.684 on the public and 0.688 on the private leaderboard respectively.

{\small
\bibliographystyle{ieee_fullname}
\bibliography{main}
}

\end{document}